\documentclass[runningheads]{llncs}

\usepackage{eccv}
\usepackage{eccvabbrv}

\usepackage{graphicx}
\usepackage{booktabs}
\usepackage{microtype}
\usepackage{multirow}
\usepackage{bm}

\usepackage{xcolor}
\usepackage{soul}
\usepackage{pifont}
\usepackage{caption}

\newcommand{\psp}{\kern0.2ex}
\newcommand{\nsp}{\kern-0.1ex}

\newcommand{\cmark}{\textcolor{green!80!black}{\ding{51}}}
\newcommand{\xmark}{\textcolor{red}{\Large$\times$}}

\usepackage[accsupp]{axessibility} 
\usepackage{orcidlink}
\usepackage{verbatim}
\usepackage{wrapfig}
\usepackage[most]{tcolorbox}

\usepackage{hyperref}

\begin{document}

\title{
Bridging Online and Offline Handwriting\\
via Differentiable Physical Rendering
}

\authorrunning{S. Park et al.}
\titlerunning{OnOff Handwriting}

\author{Seonmi Park\inst{1}\orcidlink{0009-0007-8890-554X} \and
Seunghyun Shin\inst{1}\orcidlink{0009-0006-3012-9675} \and
Vihaan Misra\inst{2}\orcidlink{0009-0002-8775-9046} \and
Dongmin Shin\inst{3}\orcidlink{0009-0003-4166-4399} \and\\
Ukcheol Shin\inst{4}\thanks{Corresponding author}\orcidlink{0000-0001-8363-9886} \and
Jean Oh\inst{2}\orcidlink{0000-0001-9709-2658} \and
Hae-Gon Jeon\inst{3}$^{\star}$\orcidlink{0000-0003-1105-1666}
}

\institute{
GIST\inst{1} \quad CMU\inst{2} \quad Yonsei University\inst{3} \quad KENTECH\inst{4}\\
\url{https://seonmip.github.io/onoff/}
}
\maketitle
\vspace{-0.35in}
\begin{figure}[h]
\centering
\includegraphics[width=\linewidth]{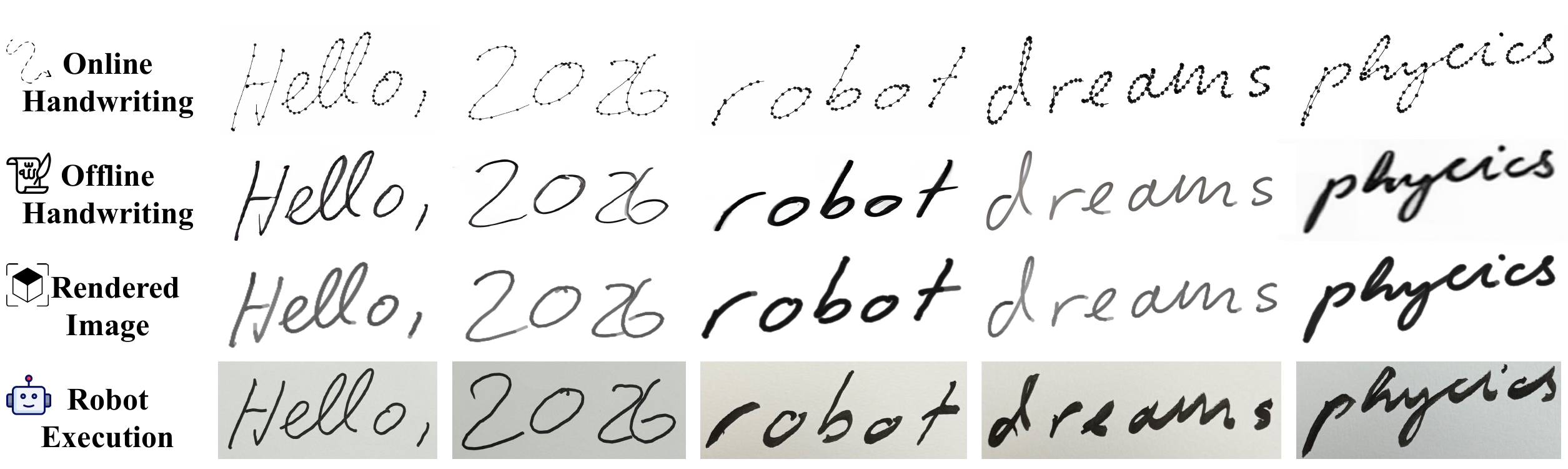}
\vspace{-0.25in}
\caption{\textbf{A unified framework for online and offline handwriting generation.} Our differentiable physical renderer connects stroke trajectories and images, enabling online handwriting trajectory estimation, offline handwriting image rendering, offline handwriting image refinement, and direct robot execution within a single framework.}
\label{fig:teaser}
\end{figure}
\vspace{-0.4in}
\begin{abstract}
Realistic handwritten text generation plays an important role in numerous applications, such as font design, biometric authentication, and robotic calligraphy.
Existing methods are typically divided into two independent paradigms: online approaches that estimate handwriting trajectories and offline approaches that synthesize realistic handwriting images. 
While online models capture structural and temporal dynamics, they often lack fine-grained textures, whereas offline models reproduce realistic appearance but discard stroke order.
However, unifying online and offline models remains challenging due to (1) the lack of an explicit physical model linking stroke kinematics to pixel-level appearance and (2) the absence of paired trajectory–image datasets.
Moreover, enabling end-to-end learning requires a differentiable rendering process across motion and appearance domains.
To address these challenges, we propose a compact physical brush model that bridges stroke dynamics and visual appearance, together with a differentiable rendering module that converts stroke trajectories into stylized images. 
By integrating these components, we propose a unified online–offline handwriting generation framework via differentiable brush rendering.
The proposed framework consists of four core modules: 1) a text-to-stroke generator that predicts the target stroke conditioned on the given text and style image, 2) a brush parameter observer that extracts brush model parameters from style references, 3) a differentiable brush renderer that maps a stroke sequence and physical brush parameters into a handwritten image, and 4) a zero-shot image refiner that refines rendered images via diffusion models. 
Extensive experiments and real-world robotic calligraphy demonstrations validate our approach, achieving both structural and visual fidelity.
\vspace{-0.1in}
\end{abstract}

\section{Introduction}
\label{sec:intro}

Generating realistic handwritten text is a cornerstone task in numerous practical applications, including artistic expression, graphics, security, robotics, and assistive technologies~\cite{berio2022strokestyles, hou2026teachingbot}. 
For example, it enables the creation of personalized fonts, the synthesis of large-scale training data for Optical Character Recognition (OCR) systems, and the development of assistive technologies for individuals with physical impairments who cannot write by hand.

More recently, it has played a key role in developing physical AI applications, such as robotic calligraphy systems that write handwritten text by a robot manipulator with a brush or pen~\cite{luo2024callirewrite}.
To support these diverse requirements, handwriting generation has traditionally been studied under two independent paradigms: online and offline.

Online handwriting generation~\cite{dai2023disentangling,liu2024elegantly,graves2013generating,tang2021write,aksan2018deepwriting,ren2024decoupling} aims to estimate a sequential trajectory, a time-ordered sequence of pen coordinates $(x, y)$ and pen states (up/down) over time. 
By explicitly modeling the writing process, online methods accurately capture the underlying motion dynamics and naturally provide an executable motion plan, making them well-suited for robotic or physically grounded applications.
However, such trajectory-based representations often omit fine-grained visual attributes, such as stroke width, ink bleeding, texture, or shape patterns that define a unique visual style.
In contrast, offline handwriting generation~\cite{dai2024one,pippi2023handwritten,bhunia2021handwriting,pippi2025zero,zhu2023conditional,gan2022higan+,kang2020ganwriting} directly synthesizes static pixel-level images.
The image-based representation can effectively capture rich texture details, including ink bleeding, brush patterns, slant, and stroke thickness. 
Nevertheless, by operating solely in the image space, they completely discard the temporal stroke order and the dynamic writing process, restricting their applicability to purely image-level generation without physical executability.

The clear and complementary nature of the two paradigms suggests that a joint online–offline handwriting model can bring theoretical and practical expansion for the field.
Integrating online and offline handwriting methods enables the model to leverage rich visual textures while benefiting from the precise structural and temporal information provided by handwriting trajectory sequences.
Therefore, the joint online–offline handwriting generation model can extend the application space by simultaneously producing visually realistic images and executable trajectories, supporting both image-level tasks and physically grounded applications.
Also, by learning the correspondence between online stroke kinematics and offline visual appearance, the unified framework enables cross-modal consistency, improves structural and stylistic fidelity, and provides a principled foundation for trajectory estimation from images and image synthesis from trajectories, even under partially observed or weakly paired data settings.

However, training a unified online–offline handwriting model is fundamentally challenging due to two core bottlenecks. 
First, there is a lack of explicit physical models that bridge stroke kinematics and pixel-level brush appearance, making the mapping between motion and texture highly under-constrained. 
Second, no existing dataset simultaneously provides both high-quality offline images and aligned sequential stroke trajectories, severely limiting its training process.
Furthermore, to enable end-to-end joint learning and to allow gradient backpropagation across the motion and appearance domains, the physical model that bridges stroke kinematics and brush appearance must be differentiable.

\begin{figure*}[t]
\centering
\begin{minipage}{0.4\linewidth}
\centering
\includegraphics[width=\linewidth]{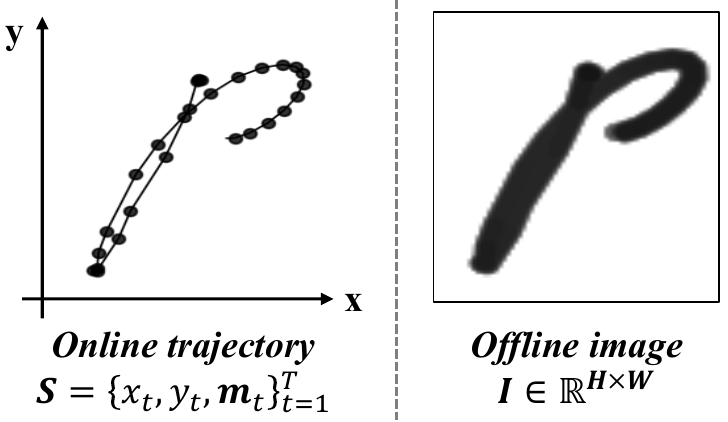}
\caption{\textbf{Illustration of online and offline handwriting.}}
\label{fig:on_offline}
\end{minipage}
\hfill
\begin{minipage}{0.58\linewidth}
\centering
\small
\captionof{table}{\textbf{Comparison of online, offline, and joint handwriting representations.}}
\label{tab:on_offline}
\vspace{0.05in}
\begin{tabular}{lccc}
\toprule
 & Online & Offline & Joint \\
\midrule
Representation & Trajectory & Image & Both \\
Structure info & \cmark & Limited & \cmark \\
Visual style & Limited & \cmark & \cmark \\
Physical execution & \cmark & \xmark & \cmark \\
Appearance realism & Limited & \cmark & \cmark \\
\bottomrule
\end{tabular}
\end{minipage}
\end{figure*}

To address the two problems, we first define a compact physical brush model that parameterizes brush behaviors using six core parameters, serving as a principled bridge between stroke kinematics and visual text appearance.
With this parameterization, we further design a differentiable rendering module that converts sequential stroke trajectories into pixel-level images.
By integrating these two key components, a physical brush model and a differentiable renderer, we propose a unified framework for joint online and offline handwriting generation.
Moreover, our proposed differentiable physical renderer enables the synthesis of stylized offline images from existing online stroke datasets, significantly expanding the applicability of trajectory-only data. 
The main contributions of this paper, whose examples are shown in~\cref{fig:teaser}, are summarized as below:
\begin{itemize}
\item We introduce a compact physical brush model with six core parameters that explicitly connect stroke kinematics to visual text appearance, providing a principled bridge between online and offline representations.
\item We design a differentiable rendering module that transforms sequential stroke trajectories into pixel-level images, enabling gradient backpropagation across motion and appearance domains.
\item We propose a unified handwriting generation framework that jointly models online trajectories and offline images via differentiable physical rendering.
\item Our renderer allows offline image synthesis from existing online stroke datasets, reducing dependency on paired datasets and broadening real-world utility.
\item We demonstrate the superior performance of our proposed method through extensive experiments and real-world robotic calligraphy demonstrations.
\end{itemize}

\section{Related Work}
\label{sec:releatedwork}
Our research builds upon two primary lines of work: handwriting generation, divided into online and offline methods and summarized in~\cref{tab:on_offline}, and physical brush modeling from the field of computer graphics.

\subsection{Handwriting Generation}
\label{sec:related_handwriting}

\textbf{Online Handwriting Generation.} 
Online handwriting generation aims to estimate an online data, a sequence of pen coordinates $(x_t,y_t)$ along with pen-up/down states, speed, or pressure information, thereby capturing the temporal dynamics of writing. 
It usually takes target text strings (\eg, "letter") as input and outputs temporal handwriting trajectory data as shown in~\cref{fig:on_offline}. This format is widely used for digital note-taking~\cite{aksan2018deepwriting} and robotic calligraphy~\cite{luo2024callirewrite}. 
The seminal work~\cite{graves2013generating} shows that recurrent neural networks can predict future pen positions point-by-point conditioned on target text strings.
Later models, such as Deepwriting~\cite{aksan2018deepwriting} and CoSE~\cite{aksan2020cose}, introduce latent disentangled models to successfully separate handwriting style from textual content, enabling editing and style transfer in trajectory space. 

Recent works have tended to focus on style-conditioned stroke sequence generation.
Write Like You~\cite{tang2021write} formulates arbitrary-style online Chinese handwriting generation as a sequence-to-sequence problem with metric-based meta learning.
SDT~\cite{dai2023disentangling} further disentangles writer-wise and character-wise style representations for stylized online character generation.
Elegantly Written~\cite{liu2024elegantly} studies online handwriting enhancement by transferring fine-grained style from a small number of user traces, while Ren \textit{et al.}~\cite{ren2024decoupling} move beyond isolated characters and address line-level online generation by decoupling layout from glyph formation.
Other works have also explored using implicit neural representations for sketches or leveraging large language models to help users write abbreviated text~\cite{xu2024skipwriter,bandyopadhyay2024sketchinr}.

\noindent\textbf{Offline Handwriting Generation.}
In~\cref{fig:on_offline}, offline handwriting generation synthesizes static pixel-level images instead of dynamic trajectories. 
It takes text strings (\eg, "letter") and style-reference images as input and estimates styled text images. 
This approach is widely used to expand datasets for text recognition models~\cite{kang2020ganwriting} and to design customized fonts for individuals~\cite{zhu2023conditional}.
Early learning-based methods like GANwriting~\cite{kang2020ganwriting} and HiGAN+~\cite{gan2022higan+} use Generative Adversarial Networks (GANs) to extract style features from a reference image and combine them to generate style-reflected text images.
Transformer-based methods strengthen style-content fusion and generalization.
HWT~\cite{bhunia2021handwriting} captures both global and local style patterns with self-attention, and Pippi \textit{et al.}~\cite{pippi2023handwritten} introduce visual archetypes to improve generalization for rare characters and unseen styles.
Recently, Diffusion Models and autoregressive architectures have been proposed to generate high-quality, arbitrary-length text lines from just a single style image~\cite{dai2024one,pippi2025zero}. 
Despite these great improvements in visual quality and diversity, offline methods cannot capture the natural stroke dynamics and writing order of human handwriting~\cite{bhunia2021handwriting,luo2024callirewrite}.

\subsection{Physical Brush Modeling and Stroke Rendering}

Our work is deeply inspired by classical research~\cite{strassmann1986hairy, lee1999simulating, chu2002efficient, baxter2001dab, chen2015wetbrush} in computer graphics that simulates the physical process of writing and drawing. 
These models provide a strong foundation for parameterizing the complex interactions among a writing tool, carried ink, and a paper surface.
Hairy Brushes~\cite{strassmann1986hairy} decomposes this process into a brush (\ie, a collection of bristles), a stroke (\ie, a trajectory of position and pressure), dip (\ie, ink application), and paper.
In that formulation, pressure affects both bristle spreading and paper contact, while overlapping marks are composed according to the darkness of the darkest stroke.
This work establishes the core idea of decoupling the stroke path from the tool's physical properties.

Subsequent research pursues higher-fidelity simulations. 
Lee~\cite{lee1999simulating} introduces soft 3D brushes whose bristle shapes vary dynamically under forces from the paper and further coupled brush rendering with paper texture and ink diffusion for oriental black-ink painting.
Chu and Tai~\cite{chu2002efficient} propose a skeleton-surface brush model solved by a constrained energy minimization, enabling brush flattening, bristle spreading, wetted-brush plasticity, and paper resistance to be reproduced in real time.
DAB~\cite{baxter2001dab} further develops a deformable 3D brush with a spring-mass skeleton and subdivision surface, together with bidirectional paint transfer and complex brush loading for interactive painting.
Wetbrush~\cite{chen2015wetbrush} pushes this line to the bristle level by jointly simulating brush, paint, and canvas interactions through coupled dynamic simulation and liquid transfer.
More recent work, InkBrush~\cite{yao2024inkbrush}, utilizes simplified geometric stroke meshes and adjustable appearance parameters to produce convincing 3D ink strokes in real time.

Although classical models can simulate brush deformation and ink transport for the writing process well, they are computationally expensive and generally incompatible with gradient-based learning, as they are neither differentiable nor easily invertible.
Therefore, our goal is not to create a perfect physical simulator but to design an inverse-friendly and differentiable model of the writing process.

\section{Differentiable Brush Renderer $\mathcal{R}$}

Our framework aims to jointly infer an online stroke sequence and generate an offline handwriting image through physically interpretable rendering parameters.
We first introduce the differentiable renderer $\mathcal{R}$, a key component of the overall framework shown in Fig.~\ref{fig:pipeline}. 
Given an online stroke sequence $S$ and brush parameters $\theta$, $\mathcal{R}$ produces a raster handwriting image $I_{\mathrm{rend}}$. 
We present $\mathcal{R}$ first because it is used to synthesize paired online--offline training data and later serves as the image formation module in the full pipeline.

\label{sec:differentiable_render}
\subsection{Brush Parameterization}
Our renderer is motivated by classical physical brush models, but is not a direct implementation of them. Instead, we distill the physical properties that are most relevant to handwriting appearance into a low-dimensional differentiable surrogate.
Chu and Tai~\cite{chu2002efficient} identify several factors that are critical for rendering visually plausible strokes. 
These include the footprint of the tool tip on the paper, pressure-dependent deformation, ink spreading under contact, and the state of the ink supply. 
Here, the footprint refers to the contact area formed between the tip and the paper surface.
We compress these factors as brush model parameters:
\begin{equation}
\theta=
\{w_{\mathrm{base}},\,k_{\mathrm{spread}},\,\rho_{\mathrm{ink}},
\,\sigma_{\mathrm{sharp}},\,p_{\mathrm{min}},\,p_{\mathrm{max}}\},
\end{equation}
where $w_{\mathrm{base}}$ denotes the base stroke width, $k_{\mathrm{spread}}$ controls the sensitivity of the width to applied pressure, $\rho_{\mathrm{ink}}$ determines the visual density of the deposited ink, and $\sigma_{\mathrm{sharp}}$ controls the boundary falloff of the stroke footprint.
These four parameters control the visible stroke appearance, including stroke width, spreading, ink intensity, and edge softness. The remaining parameters, $p_{\min}$ and $p_{\max}$, define the tool-specific dynamic range of the velocity-derived pressure proxy used in the rendering process.
Since standard online handwriting datasets typically lack physical pressure measurements, we employ a kinematic proxy estimated from the stroke's drawing velocity, assuming an inverse relationship between speed and applied pressure. 
This abstraction deliberately omits full 3D brush dynamics, wetness diffusion, and paper absorption, retaining only factors that are visually important and identifiable from style images.

\begin{figure}[t]
\centering
\includegraphics[width=\linewidth, trim={0 0 0 10.2cm}, clip]{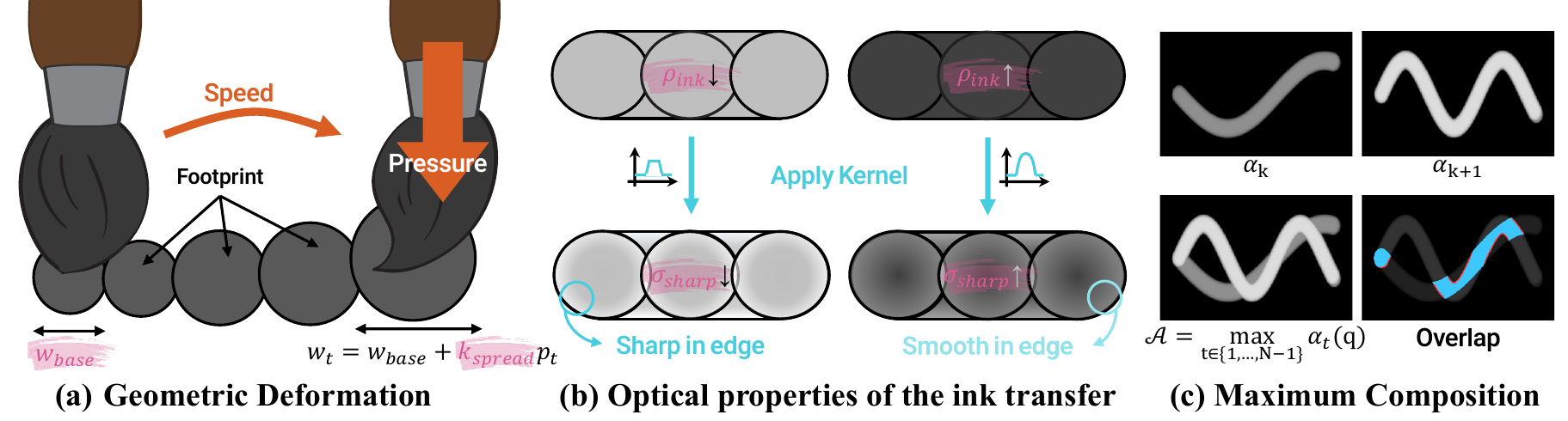}
\vspace{-7mm}
\caption{\textbf{Conceptual overview of the proposed brush model.}
(a) $w_{\mathrm{base}}$ and $k_{\mathrm{spread}}$ determine the brush footprint
geometry, with the pressure proxy bounded by $p_{\min}$ and $p_{\max}$.
(b) $\rho_{\mathrm{ink}}$ controls stroke intensity, while
$\sigma_{\mathrm{sharp}}$ controls boundary falloff.
(c) Per-step alpha maps are combined using max composition.}
\label{fig:renderer}
\vspace{-0.2in}
\end{figure}

\subsection{Differentiable Rendering Pipeline}
Our differentiable brush renderer $\mathcal{R}$ converts an online handwriting trajectory $S=\{(\mathbf{u}_t,\mathbf{m}_t)\}_{t=1}^{N}$, where $\mathbf{u}_t=(x_t,y_t)\in\mathbb{R}^2$ is the 2D pen position and $\mathbf{m}_t \in \{0, 1\}^3$ denotes the one-hot pen-state indicating pen-down, pen-up, and end-of-sequence at time step $t$, with $\theta$ into a rendered style-reflected image $I_{rend}$. The whole pipeline is depicted in Fig.~\ref{fig:renderer}. 

\noindent\textbf{Pressure proxy.}
Since online handwriting datasets do not provide pen pressure, we approximate it from the local writing speed.
Specifically, for the $t$-th stroke segment connecting consecutive pen positions $\mathbf{u}_t$ and $\mathbf{u}_{t+1}$, we compute the local writing speed as
\begin{equation}
v_t=\|\mathbf{u}_{t+1}-\mathbf{u}_t\|_2,
\qquad t=1,\ldots,N-1.
\end{equation}

and convert its stroke-wise normalized version $\tilde{v}_t\in[0,1]$ into an instantaneous pressure proxy:
\begin{equation}
p_t^{\text{proxy}}=p_{\min}+(p_{\max}-p_{\min})(1-\tilde{v}_t),
\end{equation}
where $p_{\min}, p_{\max} \in \theta$ are predicted lower and upper bounds that capture the physical response range of the writing tool.
To obtain a temporally consistent pressure signal for rendering, we apply exponential smoothing:
\begin{equation}
p_t=(1-\beta)p_{t-1}+\beta p_t^{\text{proxy}},
\end{equation}
where $\beta\in(0,1)$ controls the smoothing strength.
We can observe that pressure decreases as speed increases.

\noindent\textbf{Footprint and ink transfer.}
Classical brush models represent a stroke as a trajectory of position and pressure whose footprint varies with pressure.
Hairy Brushes further notes that increasing pressure affects both spreading and paper contact, and that spreading is taken to be linearly proportional to pressure by default~\cite{strassmann1986hairy}.
We therefore approximate the effective stroke width as follows:
\begin{equation}
w_t = w_{\mathrm{base}} + k_{\mathrm{spread}} p_t.
\end{equation}
For the segment $(\mathbf{u}_t,\mathbf{u}_{t+1})$, let $d_t(q)$ be the shortest Euclidean distance from pixel $q$ to the segment.
We define the normalized distance as:
\begin{equation}
\hat{d}_t(q)=\frac{d_t(q)}{w_t/2},
\end{equation}
and a footprint kernel as:
\begin{equation}
K_t(q)=\max\!\left(0,1-\hat{d}_t(q)^\gamma\right)\quad\text{s.t.}\quad
\gamma=\gamma_{min}+\lambda(1-\sigma_{\mathrm{sharp}}),
\label{eq:gamma}
\end{equation}
where $\gamma$ controls how sharply the response decays away from the stroke centerline.
Visible ink transfer is modeled as follows:
\begin{equation}
O_t = 1-\exp(-\rho_{\mathrm{ink}} p_t),\qquad
\alpha_t(q)=s_t\,K_t(q)\,O_t,
\end{equation}
where $O_t$ is the opacity transferred at rendering step $t$, $s_t=m_{t,\mathrm{down}}\in\{0,1\}$ is the pen-down indicator extracted from the one-hot pen state $\mathbf{m}_t$, and $\alpha_t(q)$ denotes the ink contribution of the $t$-th stroke segment at pixel $q$.

\noindent\textbf{Rendering and composition.}
The renderer outputs a continuous coverage map as below:
\begin{equation}
\mathcal{A}(q)=\max_{t\in \{1,\ldots,N-1\}}\alpha_t(q),
\end{equation}
and the grayscale rendering is defined as:
\begin{equation}
I_{\mathrm{rend}}(q)=1-\mathcal{A}(q),
\end{equation}
where $\mathcal{A}$ denotes the accumulated coverage.
We use a maximum composition rather than additive blending to avoid over-darkening in self-overlapping strokes and to obtain a stable differentiable surrogate for brush-mark accumulation.

\begin{figure}[tb]
\centering
\small
\includegraphics[width=\linewidth]{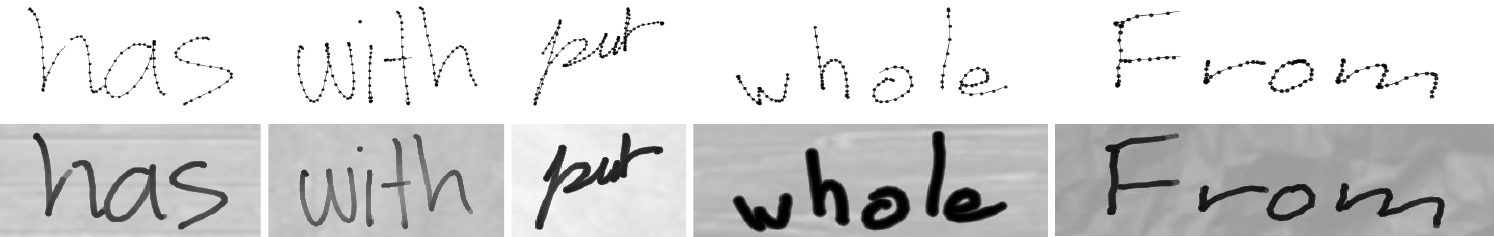}\\
\scriptsize{(a) Online stroke trajectories (top), synthetic offline image generated by our renderer (bottom).} \\ 
\vspace{0.05in}
\includegraphics[width=\linewidth]{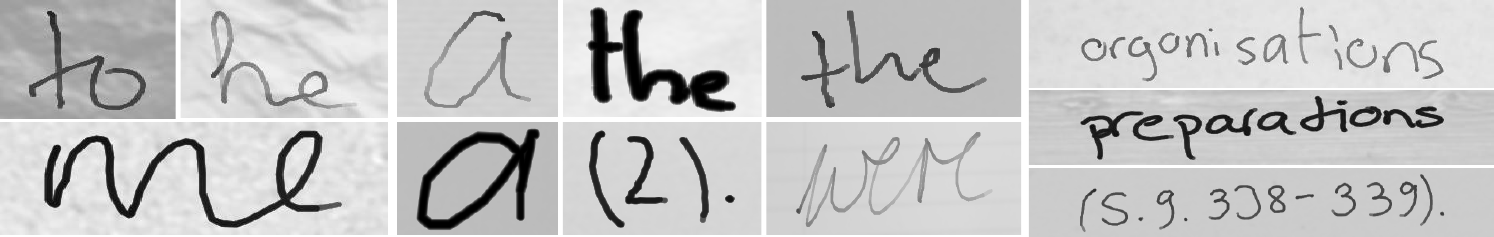}\\
\scriptsize{(b) Additional generated samples showing diverse backgrounds, styles, and word lengths} 
\vspace{-0.05in}
\caption{\textbf{Online-Offline paired dataset generated by our differentiable renderer.}}
\label{fig:dataset}
\end{figure}

\subsection{Online-Offline Paired Data Construction via Brush Rendering}
\label{sec:syn}

A primary challenge in joint online–offline handwriting modeling is the scarcity of paired datasets that provide both raster handwriting images and their corresponding stroke trajectories.
Existing online datasets only contain trajectory sequences, while offline datasets provide static images without any stroke dynamic.
To overcome this limitation, we construct a realistic paired synthetic dataset $\mathcal{D}_{\mathrm{syn}}$ from an online handwriting dataset with our differentiable brush renderer.

Let $\mathcal{D}_{\mathrm{on}}=\{(T,S)\}$ denote an online handwriting dataset consisting of text strings $T$ and their corresponding stroke trajectories $S$.
For each data sample, we randomly assign a set of physical brush parameters $\theta $ from our predefined distributions.
Using the coverage computation of $\mathcal{R}$, the online stroke sequence $S$ and
brush parameters $\theta$ are converted into a continuous coverage map
$\mathcal{A}$, which represents the spatial distribution of deposited ink.

To generate more realistic offline handwriting images, we additionally incorporate a background composition strategy inspired by Emuru~\cite{pippi2025zero}. 
We curate a collection of real-world background textures $\mathcal{B}$, including notebook paper, cardboard surfaces, and wooden boards. 
For each synthesis, a background patch $B$ is randomly sampled from $\mathcal{B}$ with random location and orientation. 
The final synthetic image $I_{\mathrm{syn}}$ is then obtained by compositing the ink onto the sampled background as follows:
\begin{equation}
I_{\mathrm{syn}} = \mathcal{A} \cdot C_{\mathrm{ink}} + (1 - \mathcal{A}) \cdot B,
\end{equation}
where $C_{\mathrm{ink}}$ denotes the ink color (e.g., black). 
This composition introduces realistic background textures and appearance variations commonly observed in offline handwriting images, allowing the model to learn complex intensity variations and noise inherent in real-world offline images.

The resulting dataset $\mathcal{D}_{\mathrm{syn}} = \{(T, S, I_{\mathrm{syn}}, \theta)\}$ provides paired supervision across both online and offline domains: (i) $(I_{\mathrm{syn}},S)$ enables the stroke generator $G$ to learn trajectory structures conditioned on visual style, and (ii) $(I_{\mathrm{syn}},\theta)$ allows the brush parameter observer $O$ to recover physically interpretable rendering parameters from appearance. The examples are shown in Fig.~\ref{fig:dataset}.

\section{Unified Online and Offline Handwriting Framework}
\label{sec:method}

\subsection{Framework Overview}
\label{sec:problem_and_overview}

\begin{figure}[tb]
\centering
\includegraphics[width=\linewidth]{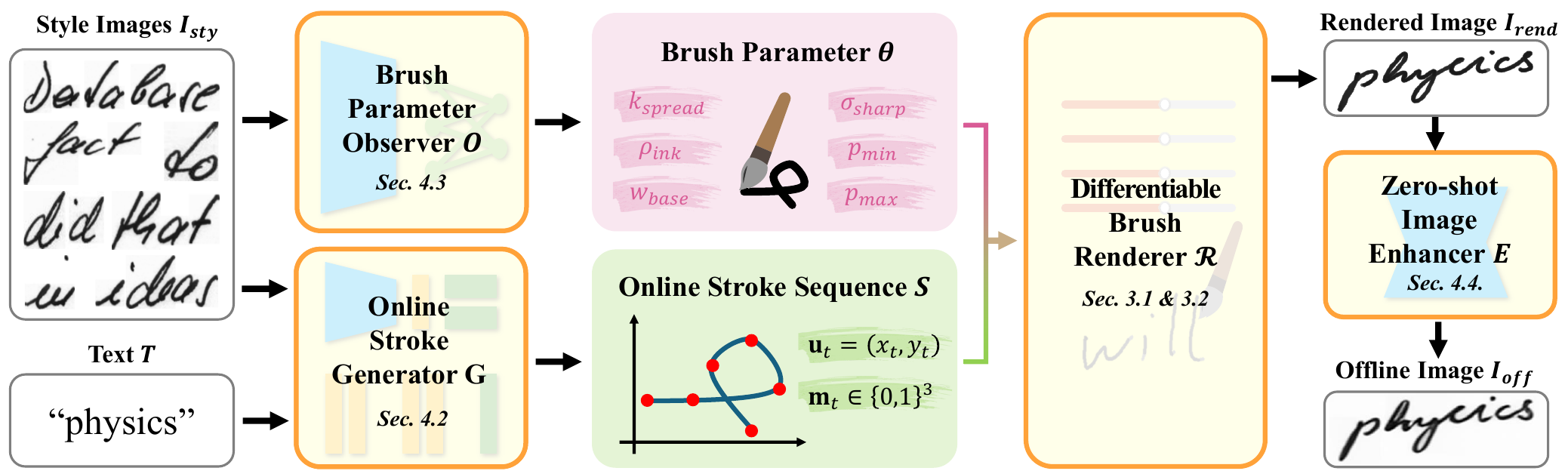}
\vspace{-5mm}
\caption{\textbf{Overview of the proposed joint online–offline handwriting framework.} Given style reference images $I_{sty}$ and a target text $T$, the brush parameter observer $O$ estimates physics-aware brush parameters $\theta$, while the stroke generator $G$ predicts an online stroke sequence $S$. A differentiable brush renderer $\mathcal{R}$ converts the online stroke sequence and brush parameters into a rendered handwriting image $I_{\mathrm{rend}}$. Finally, a zero-shot image enhancer $E$ refines the rendered result to produce the final offline handwriting image $I_{off}$ with a realistic appearance.
}
\vspace{-5mm}
\label{fig:pipeline}
\end{figure}

Given a set of handwriting style reference images $I_{sty}$ and a target text string $T=\{c_i\}_{i=1}^{L}$ where $c_i$ is a character, the brush parameter observer $O$ first estimates a set of physical brush parameters $\theta$, which encode the physical characteristics of the writing style, from the style references $I_{sty}$.
Conditioned on the text input $T$ and style references $I_{sty}$, the stroke generator $G$ predicts an online stroke sequence $S=\{(\mathbf{u}_t,\mathbf{m}_t)\}_{t=1}^{N}$.
After that, the renderer $\mathcal{R}$ converts a stroke sequence $S$ and brush parameters $\theta$ to a rendered handwriting image $I_{rend}$ in a differentiable manner.
Lastly, the image enhancer $E$ further refines the rendered output $I_{rend}$ via off-the-shelf diffusion models, which is optional.

Therefore, the overall inference process can be factorized as follows:
\begin{equation}
S = G(I_{sty}, T), \,\,\theta = O(I_{sty}), \,\, I_{\mathrm{rend}} = \mathcal{R}(S,\theta), \,\,I_{off} = E(I_{\mathrm{rend}}, I_{sty}, T),
\end{equation}
where given a set of style references $I_{sty}$ and a target text string $T$, our framework can estimate (i) an online stroke sequence $S$, (ii) a rendered image $I_{rend}$ and (iii) a refined image $I_{off}$.

\subsection{Online Stroke Generator G}
\label{sec:stroke}

The role of the online stroke generator $G$ is to produce a temporally ordered trajectory $S$ that defines the structural layout of the handwritten word.
To account for this sequential writing process, we build $G$ upon the transformer-based trajectory estimation architecture~\cite{dai2023disentangling}.
However, the original model~\cite{dai2023disentangling} is designed for single-character synthesis, which limits its applicability to practical handwriting generation scenarios where words or full text lines are typically produced.
We thus extend it to operate at the word level (\ie, multiple characters) rather than single characters.

Specifically, we replace the original single-character content token with a sequence of character embeddings derived from the target text string $T=\{c_i\}_{i=1}^{L}$, enabling the model to process multi-character inputs.
To capture contextual dependencies across characters, we introduce an additional cross-attention layer in the decoder.
This layer allows the trajectory representation to dynamically attend to the entire character sequence during generation.

With the above modifications, the generator $G$ operates at the word level and predicts the corresponding stroke sequence $S=\{(\mathbf{u}_t,\mathbf{m}_t)\}_{t=1}^{N}$ from the target text string $T$ and style images $I_{sty}$ as follows:
\begin{equation}
S=G(I_{sty}, T).
\end{equation}

\noindent\textbf{Loss function.}
The stroke generator $G$ predicts an online stroke sequence autoregressively.
At each time step, it outputs a $K$-component 2D Gaussian mixture for the pen
position and logits for the pen state, where $K=20$. We train $G$ using the
negative log-likelihood of the ground-truth pen position under the predicted
mixture distribution and a cross-entropy loss for the pen state:
\begin{equation}
\mathcal{L}_{G}
=
-\sum_{t}
\log p(\mathbf{u}_t)
+
\lambda_{\mathrm{pen}}
\sum_{t}
\mathrm{CE}(\hat{\mathbf{m}}_t,\mathbf{m}_t),
\end{equation}
where $p(\mathbf{u}_t)$ is the predicted Gaussian mixture density,
$\hat{\mathbf{m}}_t$ denotes the predicted pen-state logits, and
$\mathbf{m}_t$ denotes the ground-truth pen state. Further details are provided
in the Supplementary Material.

\subsection{Brush Parameter Observer $O$}
\label{sec:observer}
The brush parameter observer $O$ predicts brush parameters $\theta$, which controls how an online stroke sequence is rendered into a stylistic image. 
Given style reference images $I_{sty}$, the observer estimates the corresponding parameters as:
\begin{equation}
\hat{\theta} = O(I_{sty}).
\end{equation}
The estimated parameters $\theta$ can provide two functionalities: controllability and explainability. First, they control the renderer $\mathcal{R}$ to generate a style-reflected image with the handwriting trajectory. Second, they provide an interpretable description of the writing style that can be exploited for downstream systems, such as robotic calligraphy.

\noindent\textbf{Architecture.}
To capture both image-specific cues and the shared writing style, $O$ utilizes DINOv3~\cite{simeoni2025dinov3} to extract features from the reference images. The extracted features are utilized as tokens, with a learnable style token added as the first token for each reference image. The concatenated tokens are processed by an aggregator consisting of Transformer layers. The aggregator combines two approaches: intra-image self-attention, which captures local stylistic details within each individual image, and inter-image attention, which extracts a consistent shared writing style across multiple style references. The final representation of the style token is then projected through an MLP head to predict the normalized brush parameters.

\noindent\textbf{Loss function.}
We optimize $O$ using parameter-space supervision and visual rendering consistency through our differentiable brush renderer $\mathcal{R}$:
$$ \mathcal{L} = \mathcal{L}_{params} + \mathcal{L}_{render}. $$
The loss functions $\mathcal{L}_{params}=\textbf{MSE}(\hat{\theta}, \theta^{\mathrm{gt}})$ and $\mathcal{L}_{render}=\textbf{MSE}(\hat{I}_{rend}, I_{rend}^{\mathrm{gt}})$ ensure direct alignment and accurate reconstruction of the final visual appearance. 
Here, $\hat{I}_{\mathrm{rend}}=\mathcal{R}(S,\hat{\theta})$ and $I_{\mathrm{rend}}^{\mathrm{gt}}=\mathcal{R}(S,\theta^{\mathrm{gt}})$.

\subsection{Zero-shot Offline Image Enhancer $E$}
\label{sec:offline}
The rendered image $I_{\mathrm{rend}}$ effectively captures the desired word structure and a style-reflected brush footprint.
However, several real-world handwriting effects, commonly observed in offline images, remain beyond the scope of the renderer, such as paper texture, scanner artifacts, local intensity fluctuations and dataset-specific appearance statistics.
To mitigate this gap, we employ an off-the-shelf handwriting diffusion model~\cite{dai2024one, nikolaidou2024diffusionpen} as a zero-shot image enhancer $E$.

A key design choice is to keep the diffusion model completely unchanged.
Rather than retraining the model or modifying its architecture, we utilize the rendered image $I_{rend}$ as a structural prior during the diffusion sampling process.
This strategy allows us to exploit a pretrained handwriting diffusion model in a zero-shot manner while preserving the stroke structure predicted from our framework.

Let $\mathcal{E}$ denote the VAE encoder of the diffusion model.
We first encode the rendered image into the latent space to obtain a structural prior $z_{\mathrm{rend}}$:
\begin{equation}
z_{\mathrm{rend}}=\mathcal{E}(I_{\mathrm{rend}}).
\end{equation}
At a chosen intermediate timestep $t_0$, we initialize the diffusion latent as:
\begin{equation}
z_{t_0} = \sqrt{\bar{\alpha}_{t_0}}\,z_{\mathrm{rend}} + \sqrt{1-\bar{\alpha}_{t_0}}\,\epsilon
\quad\text{s.t.}\quad \epsilon\sim\mathcal{N}(0,I),
\end{equation}
and then run the standard conditional denoising process conditioned on $(T,I_{sty})$ to estimate the denoised latent $z_0$.
Starting the diffusion process from a noised version of the rendered structure rather than pure noise preserves the generated layout while allowing the model to refine the appearance.
Lastly, the refined offline handwriting image $I_{off}$ is obtained by decoding the denoised latent $z_0$ with the VAE decoder $\mathcal{D}$:
\begin{equation}
I_{off}=\mathcal{D}(z_0).
\end{equation}

\section{Experiments}
\label{sec:experiments}

Here, we describe the training and evaluation datasets used for our framework. 
Please refer to the Supplementary Material for implementation details, evaluation metrics, and further experimental settings.

\noindent\textbf{Training: online-offline paired dataset.}
For training our proposed framework, we construct an online–offline paired dataset by combining IAM-OnDB~\cite{liwicki2005iam} and CASIA-OLHWDB~\cite{liu2011casia} with the data construction pipeline that utilizes our differentiable renderer (\ie, Sec.~\ref{sec:syn}).
For each stroke sequence $S$ and text string $T$, we sample brush parameters $\theta$ and render a corresponding handwriting image $I_{\mathrm{syn}}$, forming synthetic pairs $(T, S, I_{\mathrm{syn}}, \theta)$ used for training the stroke generator and brush parameter observer.
The resulting dataset contains 155,840 word-level samples, split by writer into 303 writers for training and 78 for testing (\ie, online handwriting evaluation).

\noindent\textbf{Evaluation: Online handwriting dataset.}
We utilize IAM-OnDB~\cite{liwicki2005iam} and CASIA-OLHWDB (1.0--1.2)~\cite{liu2011casia} to generate a paired dataset and evaluate online handwriting results.
Each trajectory contains on average $\sim$60 points and is truncated to a maximum length of 250 points.
The maximum word length is 9 characters with an average of 3.5, and the dataset includes 25,912 unique word types.
Since IAM-OnDB provides line-level annotations, we segment each line into word-level samples using the corresponding labels and retain only words verified through an OCR-based filtering step (see Supplementary Material for details).

\noindent\textbf{Evaluation: Offline handwriting datasets.}
For offline handwriting evaluation, we use the test sets of IAM~\cite{marti2002iam} and CVL~\cite{kleber2013cvl}, containing 161 and 283 writers respectively.
Words longer than 9 characters are discarded.
All rendered images are resized to a fixed height of 64 pixels while preserving the original aspect ratio.

\subsection{Evaluation Results}
\noindent\textbf{Quantitative Comparison}
For evaluation, we utilize a test set comprising IAM-OnDB~\cite{liwicki2005iam} and CASIA~\cite{liu2011casia}, along with their rendered images. The performance is assessed in both single-letter and multi-letter scenarios. In the multi-letter setting, since SDT~\cite{dai2023disentangling} is designed for single-letter online handwriting, we perform inference on individual characters and concatenate them for a fair comparison. As shown in Table~\ref{tab:online_eval}, our stroke generator outperforms the state-of-the-art single-character model in terms of quality, demonstrating the robustness and scalability of our proposed method



In table~\ref{tab:syn_eval}, comparative analysis reveals that standard offline handwriting models struggle to generalize when conditioned on our paired dataset, primarily due to an over-reliance on pixel-level distributions. Since our dataset utilizes real human trajectories, the failure of these baselines highlights their inability to capture the underlying motion. Our model overcomes this by explicitly decoupling and modeling the stroke-level physics; the performance gains confirm that our Stroke Generator effectively bridges the gap between digital trajectories and realistic image synthesis.


For Ours variants, the noise-injection start step is tuned for each baseline--dataset setting: 50 for DiffPen+Ours on both IAM and CVL, 900 for One-DM+Ours on IAM, and 200 for One-DM+Ours on CVL.
Table~\ref{tab:offline_eval} summarizes the quantitative comparison on IAM Words and CVL Words.
Among the baseline models, One-DM~\cite{dai2024one} shows the strongest overall performance, achieving the best FID, BFID, and CER on both datasets, while DiffPen~\cite{nikolaidou2024diffusionpen} yields the lowest HWD~\cite{pippi2023hwd} on both datasets.
By contrast, Emuru~\cite{pippi2025zero} performs noticeably worse across most metrics.

Across two diffusion baselines and two datasets, attaching our renderer consistently improves FID and BFID, showing that the proposed prerender-guided initialization improves visual realism. 
Other metrics, such as HWD and CER, show backbone- and dataset-dependent trade-offs, indicating that the optimal noise-injection start step should be tuned for each setting rather than fixed globally.
Overall, our framework acts as a complementary guidance module whose effect depends on the backbone model, primarily improving realism-oriented metrics while improving handwriting-style consistency in some settings.

\begin{table*}[t]
    \centering
    \begin{minipage}{0.28\linewidth}
        \centering
        \centering
\caption{Evaluation of online handwriting models in two scenarios.}
\vspace{-2mm}
\label{tab:online_eval}
\resizebox{\linewidth}{!}{
\begin{tabular}{lcc}
\toprule
& \multicolumn{2}{c}{\textbf{DTW $\downarrow$}} \\
\cmidrule{2-3}
\textbf{Model} & \begin{tabular}{c} multi-\\letter \end{tabular} & \begin{tabular}{c} single-\\letter \end{tabular} \\
\midrule
SDT  & 0.8155 & 0.6056 \\
Ours & \textbf{0.2936} & \textbf{0.3462} \\
\bottomrule
\end{tabular}
}
    \end{minipage}
    \hfill 
    \begin{minipage}{0.69\linewidth}
        \centering
        \centering
\caption{Evaluation on our On-Offline Paired Dataset. \textbf{Bold}: Best, \underline{Underline}: Second-best.}
\vspace{-2mm}
\label{tab:syn_eval}
\resizebox{\linewidth}{!}{
\begin{tabular}{lccccc}
\toprule
\textbf{Model} & \textbf{FID $\downarrow$} & \textbf{BFID $\downarrow$} & \textbf{HWD $\downarrow$} & \textbf{CER $\downarrow$} & \textbf{LPIPS $\downarrow$} \\
\midrule
VATr++       & 74.28 & 33.44  & 2.5990 & \underline{0.329} & 0.622 \\
DiffPen & 89.06 & 118.04 & \underline{2.0030} & 0.496 & 0.602 \\
One-DM       & \underline{52.26} & \underline{19.88}  & 2.3650 & \textbf{0.229} & \underline{0.539} \\
Emuru        & 89.97 & 60.88  & 2.6220 & 6.126 & 0.583 \\
\midrule
\textbf{Our Renderer} & \textbf{11.81} & \textbf{11.03} & \textbf{1.0833} & 0.361 & \textbf{0.001} \\
\bottomrule
\end{tabular}
}

    \end{minipage}
\end{table*}

\begin{table}[t]
\vspace{-2mm}
\caption{Quantitative evaluation for baseline models and our framework on IAM and CVL datasets. \textbf{Bold}: Best, \underline{Underline}: Second-best.}
\vspace{-3mm}
\centering\small
\resizebox{\linewidth}{!}{
\begin{tabular}{lcccccccccccc}
\toprule
\multicolumn{1}{c}{\textbf{Dataset}}
& \multicolumn{5}{c}{\textbf{IAM Words}}
& \multicolumn{5}{c}{\textbf{CVL Words}} \\
\cmidrule(lr){1-1} \cmidrule(lr){2-6} \cmidrule(lr){7-11}
\multicolumn{1}{c}{\textbf{Model}}
& FID$\downarrow$ & BFID$\downarrow$ & HWD$\downarrow$ & CER$\downarrow$ & LPIPS$\downarrow$
& FID$\downarrow$ & BFID$\downarrow$ & HWD$\downarrow$ & CER$\downarrow$ & LPIPS$\downarrow$ \\
\midrule
\textbf{VATr++}
    & 52.57 & 34.17 & 2.1668 & 0.3900 & 0.4465
    & 25.11 & 20.12 & 2.2394 & 0.4775 & 0.4034 \\
\textbf{DiffPen}
     & 77.70 & 151.96 & \textbf{1.6275} & 0.6672 & 0.5380
    & 46.31 & 108.21 & \underline{1.6057} & 0.6056 & 0.4680 \\
\textbf{One-DM}
    & \underline{29.00} & \underline{16.35} & 1.8306 & \textbf{0.3117} & \underline{0.3726}
    & \underline{19.45} & \underline{15.18} & 2.1500 & \textbf{0.2877} & \underline{0.3898} \\
\textbf{Emuru}
    & 89.89 & 72.64 & 2.2464 & 3.0583 & 0.5453
    & 68.06 & 54.52 & 2.2366 & 3.5744 & 0.4816 \\
\textbf{DiffPen+Ours}
    & 76.86 & 97.19 & 2.3116 & 0.7103 & 0.5155
    & 30.68 & 64.99 & 2.1506 & \underline{0.3918} & 0.4485 \\
\textbf{One-DM+Ours}
    & \textbf{25.89} & \textbf{8.15} & \underline{1.6442} & \textbf{0.3117} & \textbf{0.3288}
    & \textbf{14.45} & \textbf{10.47} & \textbf{1.5264} & 0.6285 & \textbf{0.3441} \\
\bottomrule
\end{tabular}
}
\label{tab:offline_eval}
\vspace{-5mm}
\end{table}

\begin{figure}[tb]
\centering
\includegraphics[width=\linewidth]{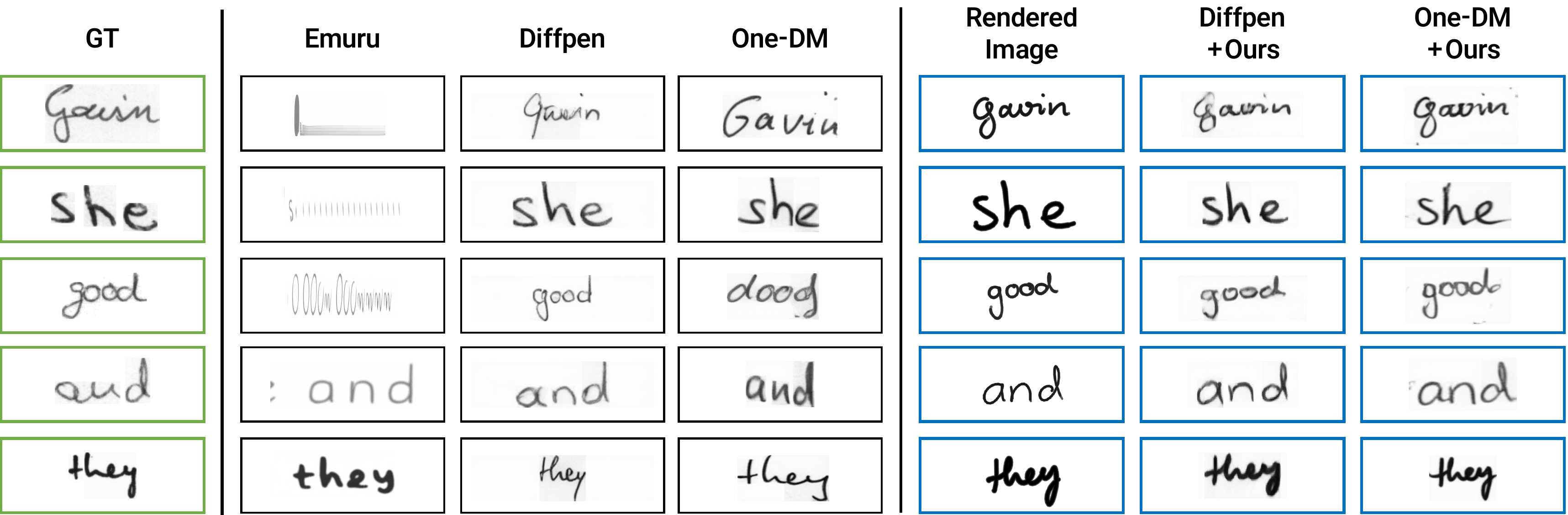}
\vspace{-5mm}
\caption{\textbf{Qualitative comparison of offline handwriting generation results.} From left to right: ground-truth images, existing offline handwriting generation methods, and our results (\ie, rendered result + zero-shot enhancer $E$).
While previous methods often fail to preserve the correct word structure or produce unstable strokes, combining them with our method improves structural consistency and visual realism.
}
\label{fig:quali_off}
\end{figure}

\noindent\textbf{Qualitative Results.}
We present results from our method and competing methods in Fig.~\ref{fig:quali_off}. 
Our approach combines a differentiable stroke renderer with a diffusion-based
sampling process. 
The renderer provides physically grounded guidance that
preserves style-relevant attributes such as stroke thickness and ink density,
while the diffusion model enhances visual realism and texture.
As a result, the generated samples remain faithful to the underlying writer-specific stroke characteristics while maintaining realistic appearance.

We further conduct a ranking-based user study for perceptual style similarity.
One-DM+Ours obtains the best average rank of 2.840, improving over One-DM with 3.406. Detailed protocols and full results are provided in the Supplementary Material.

\begin{figure}[t]
\centering
\includegraphics[width=\linewidth]{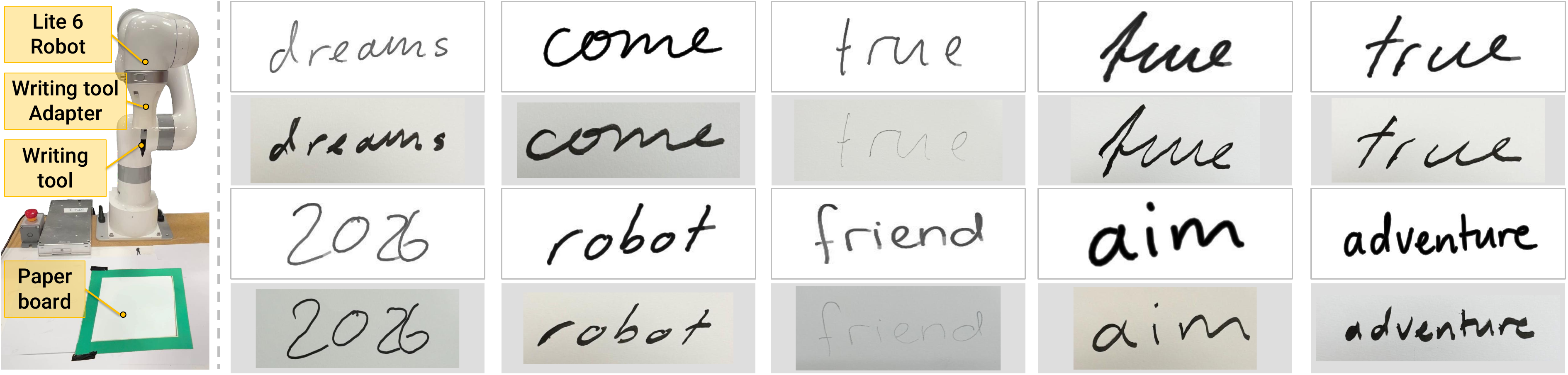}
\vspace{-7mm}
\caption{\textbf{Qualitative results of our framework executed by a real robotic system.}
For each word, the upper row shows the offline handwriting image rendered by our differentiable brush renderer, while the lower row shows the corresponding robotic writing obtained by executing the predicted stroke trajectory. 
These examples demonstrate that our method generates visually consistent handwriting while producing physically executable trajectories for robotic manipulation.
}
\label{fig:robot_demo}
\end{figure}

\subsection{Application: Robotic Writing Demonstration}
To demonstrate the physical applicability of our framework, we execute the predicted stroke trajectories $S$ using a UFACTORY Lite 6 robot arm.
Given a reference handwriting style image, our model first estimates the corresponding stroke parameters $\theta$, which characterizes properties such as base stroke width and ink density. 
Based on these predicted parameters, we select a physical writing tool (\ie, pencil, sharpie, marker pen, etc) whose tip characteristics best match the estimated stroke footprint and ink deposition behavior.
A writing tool is fixed to the robot end-effector (\ie, 0-DoF rotation), and the generated trajectory $S$ is directly executed as Cartesian motions.
The 2D stroke coordinates from $S$ define the $(x,y)$ path, while the pressure proxy is mapped to the vertical position $z$, controlling the contact depth between the pen and the surface.
This simple mapping enables the robot to reproduce the generated handwriting without additional optimization or post-processing, showing that our framework can be directly deployed in robotic writing applications, as shown in Fig.~\ref{fig:robot_demo}.

\section{Conclusion}
\label{sec:conclusion}

We introduce a compact physics-aware brush parameterization and a differentiable brush renderer that connects stroke trajectories to pixel-level handwriting images. 
With this formulation, we develop a unified framework that bridges online and offline handwriting generation, consisting of a differentiable brush renderer, an online stroke generator, a brush parameter observer and a zero-shot offline image enhancer. 
Extensive experiments demonstrate that our method can generate visually consistent handwriting images while producing physically executable trajectories that can be directly deployed on a robotic writing system. 

\noindent\textbf{Limitations.}
Although the proposed brush parameters are physically interpretable, several of them are renderer-level surrogate parameters rather than quantities directly calibrated to real-world physical units. Therefore, transferring the predicted parameters to physical writing tools may require additional calibration.
Furthermore, the current architecture is optimized for word-level generation. Extending the framework to coherent long-form sentence generation remains an important direction for future work.

\section*{Acknowledgements}
This work was supported by GIST-IREF from Gwangju Institute of Science and Technology(GIST), Korea Institute of Science and Technology (KIST) Institutional Program (26K0030) and the Institute of Information \& Communications Technology Planning \& Evaluation (IITP) grant funded by the Korean Government (MSIT) (No. RS-2024-00457882 and No.RS-2025-25441838) and the National Research Foundation of Korea(NRF) grant funded by the Korea government(MSIT)(RS-2024-00338439).

\appendix

\renewcommand{\thefigure}{\Roman{figure}}
\renewcommand{\thetable}{\Roman{table}}
\author{}
\institute{}
\title{Appendix} 
\maketitle

\section{Motivation: Handwriting as a Single Physical Act}

Handwriting is not merely pixels or coordinates; it is a single physical act. The movement of a writing instrument produces a visual trace, yet existing research typically treats online trajectories and offline images as independent generative tasks. This separation creates a causal gap between motion and appearance: image based models may generate structurally inconsistent glyphs because they are not grounded in the physical processes of writing, while trajectory only models fail to capture the material richness of the resulting visual form. To address this gap, our framework models handwriting as a unified physical process and introduces a compact brush model parameterized by $\theta$, where $\theta$ denotes the brush parameters governing stroke formation and linking trajectory generation with its resulting image.

\section{Implementation Details}

\noindent\textbf{Online Stroke Generator $G$.} 
The style encoder consists of a modified ResNet-18~\cite{he2016deep} backbone followed by a two-layer Transformer with $d_{model} = 512$ and $n_{head} = 8$. For the decoding process, we employ three-layer Transformer decoders; this includes a text decoder conditioned on multi-character text strings. The model produces a 123-dimensional output vector, which parameterizes a 20-mixture 2D Gaussian Mixture Model ($20 \times 6$ parameters) and three discrete pen states. Since the output represents the probability distribution of the subsequent stroke point, this formulation is inherently applicable to both single-character and multi-character synthesis. During inference, the stroke generator is conditioned on 15 reference style images that provide the writing style.

\vspace{3mm}
\noindent\textbf{Brush Parameter Observer $O$.}
We adopt DINOv3 (ConvNeXt Tiny)~\cite{simeoni2025dinov3} as a frozen backbone with 768-dimensional features.
Following the VGGT architecture~\cite{wang2025vggt}, we design an aggregator composed of Transformer layers that interleave frame-wise and global self-attention (8 heads).

During training, four reference style images are provided as input. At inference time, two images are randomly sampled from the fifteen style references used in the stroke generator.
All images are resized to $H=256$ with a patch size of 16.

The final style token representation is passed through a small MLP head (768$\rightarrow$512$\rightarrow$6) with Sigmoid activation to predict the brush parameters in $[0,1]$.

\vspace{3mm}
\noindent\textbf{Datasets and Evaluation.}
We construct a paired online and offline dataset using our renderer by combining IAM OnDB~\cite{liwicki2005iam} with synthetic samples and CASIA OLHWDB (1.0–1.2)~\cite{liu2011casia}, resulting in 155,840 samples in total. For IAM OnDB, we convert multi-word trajectories into single-word samples and remove samples whose rendered trajectories form straight lines using an OCR based filtering step applied to the rendered images.

Trajectories are simplified using the Ramer–Douglas–Peucker algorithm with $\epsilon = 2$ and filtered to a maximum length of $N \le 250$ (99\% of IAM samples contain fewer than 170 points). Each point is represented as a 5D vector
$[\Delta x, \Delta y, m_{1}, m_{2}, m_{3}]$.
Content is represented using a vocabulary of 73 characters consisting of punctuation symbols (! " ' ( ) , - . : ?), digits, and upper and lower case alphabet characters. All images are normalized to a fixed height of 64, with the width scaled proportionally.

\vspace{3mm}
\noindent\textbf{Training Hyperparameters.} 
Both models are optimized using Adam optimizer with a learning rate of $2 \times 10^{-4}$, a gradient clipping threshold of 5.0 ($L_2$). For the online stroke generator, we train for 200,000 iterations with 20,000 warmup steps on 6 NVIDIA RTX 3090 GPUs. The brush parameter Observer is trained for 100,000 iterations with 10,000 warmup steps on a single RTX 3090 GPU.

\section{Brush Renderer and Dataset Construction}

\subsection{Brush Renderer Details}

\begin{figure}[tb]
\centering
\includegraphics[width=\linewidth]{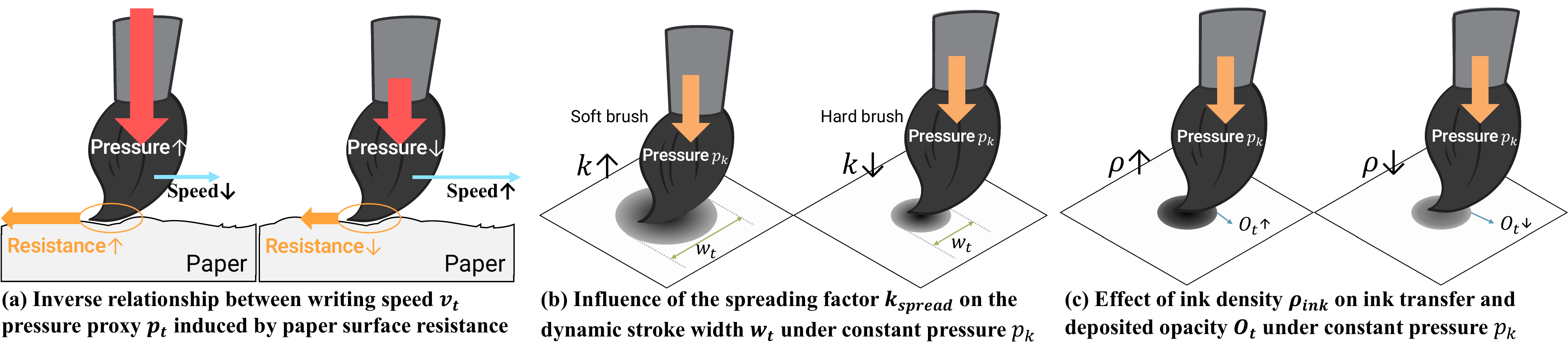}
\vspace{-5mm}
\caption{Brush parameter in the proposed differentiable brush renderer.
}
\label{fig:renderer_details}
\end{figure}

\noindent\textbf{Parameter Interpretability.}
Eq. ({\color{red}1}) encapsulates the visual identity of a writing tool into a low-dimensional vector $\theta$. As illustrated in Fig.~\ref{fig:renderer_details}, $w_{\mathrm{base}}$ and $k_{\mathrm{spread}}$ govern the geometric deformation, while $\rho_{\mathrm{ink}}$ and $\sigma_{\mathrm{sharp}}$ control the optical properties of the ink transfer, namely ink tones. This modularity allows our framework to represent diverse media, from fine-tip ballpoint pens to expressive calligraphy brushes.

\noindent\textbf{Velocity to Pressure Mapping.}
The core of our pressure proxy in Eqs. ({\color{red}2-4}) lies in the physical constraint that higher writing speeds often lead to reduced contact pressure due to the resistance of paper pores as shown in Fig.~\ref{fig:renderer_details}(a). 

\noindent\textbf{Brush Footprint and Edge Sharpness.}
Figs.~\ref{fig:renderer_details}(b) and~3(a) visualize the logic of Eq.~(5). The effective width $w_t$ expands linearly with pressure $p_t$, a simplification of the complex tuft flattening process in which bristles spread laterally upon contact. Fig.~3(b) shows Eqs.~(6--7). The parameter $\gamma$, controlled by $\sigma_{\mathrm{sharp}}$, defines the falloff profile of the footprint kernel $K_t$. A lower $\sigma_{\mathrm{sharp}}$ results in a steeper decay, mimicking a stiff, technical pen with crisp edges, while a higher value produces softer, more diffused boundaries typical of absorbent bristle brushes.

\noindent\textbf{Composition Logic.}
The final rendering step in Eqs.~(8--10) employs maximum composition. Traditional additive blending in digital rendering often leads to unrealistic over-darkening at stroke intersections. By following the principle that the darkness at an intersection is determined by the darkest overlapping stroke, our renderer maintains a stable and physically plausible appearance even in complex, self-overlapping handwriting trajectories, as shown in Fig.~\ref{fig:renderer_details}(c) and Fig.~3(c).

\subsection{Dataset Synthesis Pipeline}

\begin{figure}[tb]
    \centering
    \begin{minipage}{0.55\textwidth}
        \centering
        \includegraphics[width=\textwidth]{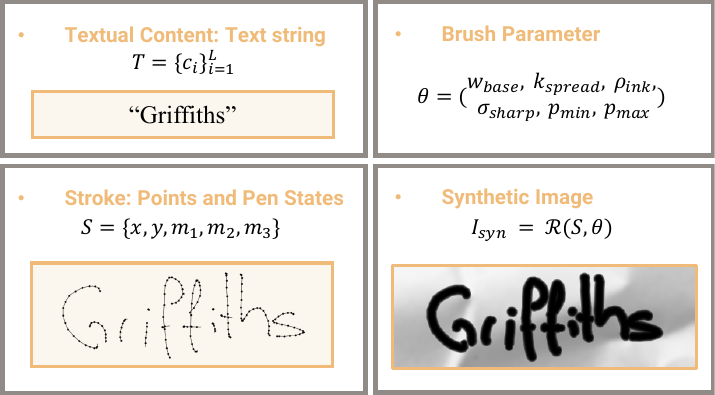}
        \caption{Composition of the Online-Offline Paired Dataset $\mathcal{D}_{\mathrm{syn}} = \{(T, S, I_{\mathrm{syn}}, \theta)\}$.}
        \label{fig:paired_dataset}
    \end{minipage}
    \hfill
    \begin{minipage}{0.41\textwidth}
        \centering
        \includegraphics[width=\textwidth]{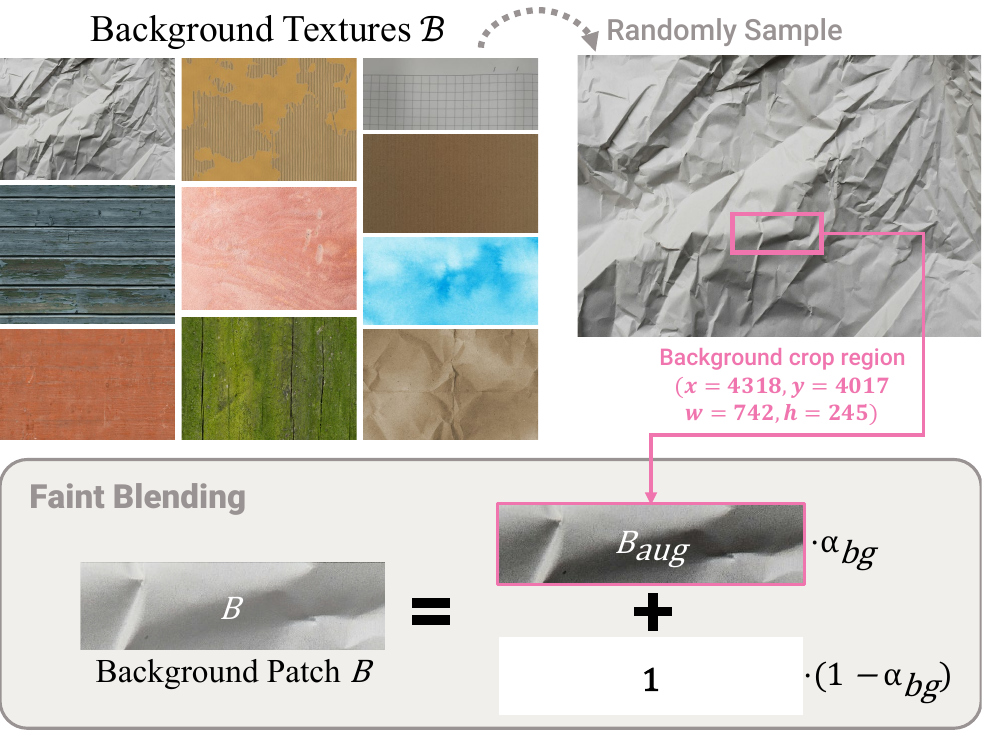}
        \caption{Background patch $B$ from background textures $\mathcal{B}$.}
        \label{fig:background}
    \end{minipage}
\end{figure}

An overview of the dataset composition is illustrated in Fig.~\ref{fig:paired_dataset}.

\noindent\textbf{Background Synthesis and Faint Blending.}
As mentioned in the main paper, we use a collection of real-world textures $\mathcal{B}$ for the background of the Online-Offline paired dataset. To simulate the natural appearance of ink absorption and paper reflection, we implement a faint blending strategy as shown in Fig.~\ref{fig:background}. Given an augmented background patch $B_{aug}$, the final canvas $B$ is generated by:

\setcounter{equation} {18}
\begin{equation}
    B = B_{aug} \cdot \alpha_{bg} + \mathbf{1} \cdot (1 - \alpha_{bg})
\end{equation}
where $\alpha_{bg}$ is randomly sampled from $\mathcal{U}(0.35, 0.4)$ and $\mathbf{1}$ denotes a white canvas. This prevents the background texture from interfering with the ink's visibility while maintaining realistic paper grains.

\noindent\textbf{Detailed Stroke Profile Control.}
The sharpness of the stroke boundary is controlled by the parameter $\sigma_{\mathrm{sharp}}$. In our implementation, the decay exponent $\gamma$ of the footprint kernel in Eq.~({\color{red}7}) of the main paper is dynamically determined using $\gamma_{\min}=2.0$ and $\lambda=8.0$. This formulation allows the renderer to represent a wide range of writing tools, from soft brush marks ($\gamma \approx 2.0$) to sharp pen-like strokes ($\gamma \approx 10.0$). 

\section{Robotic Implementation Details}

To qualitatively verify the physical executability of our differentiable renderer $\mathcal{R}$, we adopt the FRIDA framework~\cite{schaldenbrand2022frida, chen2025spline} to conduct a real-world robotic writing demonstration across diverse writing instruments.
This section details the hardware setup and the mapping logic between abstract parameters and real-world control variables.

\subsection{Robot Platform and Writing Instruments}

\noindent\textbf{Robot Platform.} We use a UFACTORY Lite 6 robot arm equipped with a writing-tool adapter that rigidly fixes the tool orientation relative to the end effector.

\noindent\textbf{Execution Framework.} 
The pen positions $\{\mathbf{u}_t\}$ from the online stroke sequence $S=\{(\mathbf{u}_t,\mathbf{m}_t)\}$ are mapped from the normalized canvas space $[0,1]^2$ to the robot's global workspace coordinates $(X,Y)_{\mathrm{global}}$ via an affine transformation $\mathcal{T}_{\mathrm{aff}}$ that accounts for scaling and aligns the canvas with the robot's base frame.

\noindent\textbf{Selected Instruments.} We selected four distinct writing tools to test the model's versatility across different physical responses:

\begin{figure}[tb]
    \centering
    \includegraphics[width=\textwidth]{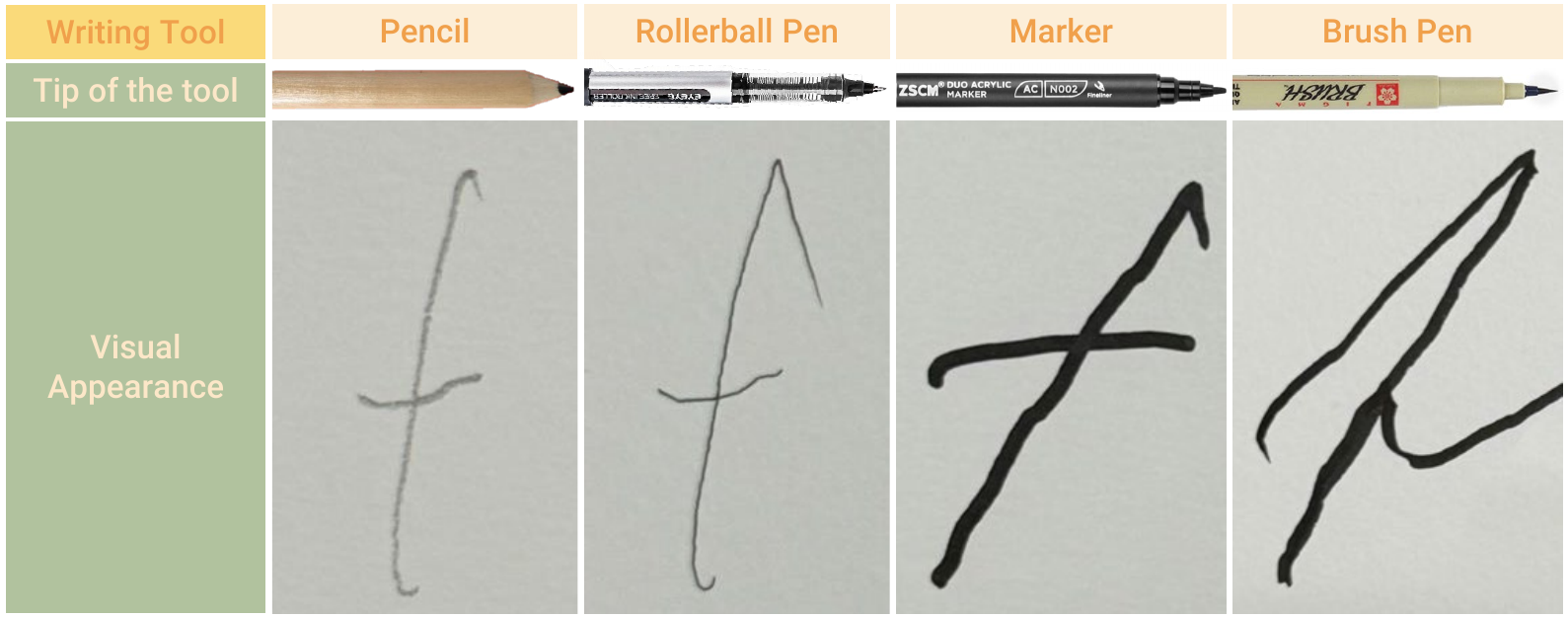}
    \vspace{-5mm}
    \caption{The shape of the tip of writing tools and its visual appearance on the canvas.}
    \label{fig:writing_tool}
    \vspace{-5mm}
\end{figure}

\begin{itemize}
    \item \textbf{Pencil}: A Jumbo Triangular Wooden Pencil (natural wood), chosen for its thick lead to test static penetration depth and subtle width variations.
    \item \textbf{Rollerball Pen}: EYEYE 0.5mm Extra Fine Point Rollerball Pen, characterized by low friction and rapid liquid ink flow, ideal for testing rapid-dry performance.
    \item \textbf{Marker}: ZSCM Dual-Tip Acrylic Ink Pen (Marker-Fineliner tip), providing opaque, consistent ink deposition with a resilient fiber tip.
    \item \textbf{Brush Pen}: Sakura Pigma Micron Brush Black Pen, featuring a flexible tip that is highly sensitive to $Z$-axis displacement, which is crucial for validating our pressure-driven spreading model, $k_{\text{spread}}$.
\end{itemize}
All writing instruments used in our experiments were purchased from Amazon.

\subsection{Parameter Mapping to Robot Control}

\begin{figure}[tb]
    \centering
    \includegraphics[width=\textwidth]{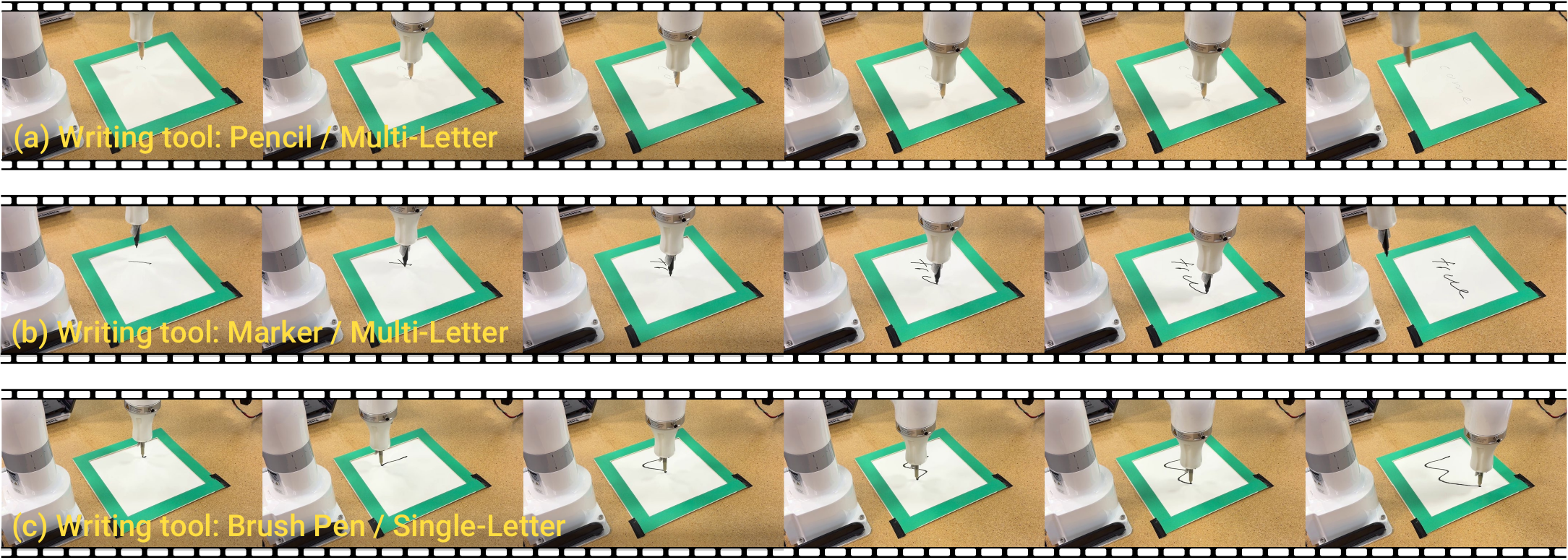}
    \caption{Qualitative demonstration of robot execution in our unified online-offline handwriting framework across diverse instruments and letter scenarios.}
    \label{fig:robot_demo_video}
\end{figure}

Examples of robotic writing execution are shown in Fig.~\ref{fig:robot_demo_video}.
We bridge the gap between our renderer $\mathcal{R}$ and physical execution by mapping the estimated parameters $\theta$ to the robot's control variables.

\noindent\textbf{Depth Control.} 
We simulate pressure by modulating the vertical penetration depth:
\begin{equation}
Z_{\text{target}} = Z_{\text{canvas}} - z_{\text{base}} - z_{\text{dynamic}}
\end{equation}
The static offset $z_{base} = w_{base} \times \lambda_{base}$ ensures consistent initial paper contact where $\lambda_{base}$ is a scaling factor mapping pixel-width to physical depth. The dynamic component $z_{dynamic} = \text{clip}(p_t \cdot k_{spread} \cdot \lambda_{dyn},\, 0,\, z_{max})$ which comes from Eq. ({\color{red} 5}) in main paper scales the depth based on the tool flexibility and intent. The clip operation prevents hardware damage and accounts for the physical saturation of the tool tip.

\noindent\textbf{Reference Plane Calibration.} 
The term $Z_{canvas}$ denotes the calibrated vertical coordinate of the physical writing surface within the robot workspace. Before execution a manual or automated calibration procedure is performed to identify the exact contact plane where the brush tip meets the paper without applying pressure. This constant serves as the reference datum. All subsequent depth-related parameters such as $z_{base}$ and $z_{dynamic}$ are subtracted from $Z_{canvas}$ to achieve the desired brush penetration and resulting stroke width.

\begin{figure}[t]
\centering
\includegraphics[width=\linewidth]{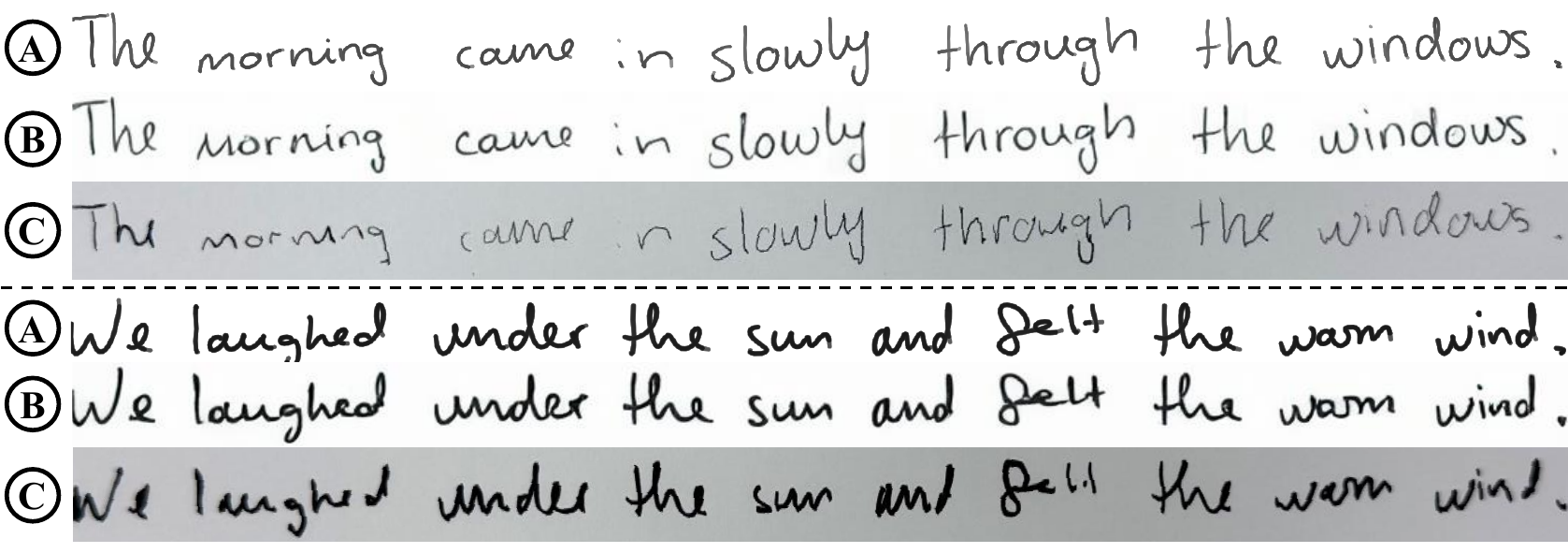}
\caption{Sentence-level application. (A) rendered image, (B) offline image, and (C) robot execution results.}
\vspace{-5mm}
\label{fig:sentence}
\end{figure}

\noindent\textbf{Scope and Scalability beyond Word-level Generation.}
Our differentiable renderer is not restricted to word-level generation. It operates on arbitrary stroke sequences. We can easily compose word-level trajectories into sentence-level sequences by adding a fixed spatial offset to each word's trajectory. We then use this combined trajectory for offline generation guidance and robotic execution, as shown in Fig.~\ref{fig:sentence}.
This demonstrates that our pipeline is scalable beyond the word level.
Nevertheless, the fully end-to-end sentence generation remains our future work. To do this, we have to overcome the issue of the paired data dependency. One of the potential solutions is leveraging the sequence-level online handwriting dataset~\cite{liwicki2005iam} by converting it into our data format to retrain our framework.

\section{Extended Evaluation and Analysis}

\subsection{Details of Evaluation Metrics}

We evaluate the quality of generated handwriting in both online (trajectory) and offline (image) domains. 
For online handwriting trajectories, we use \textbf{Dynamic Time Warping (DTW)}~\cite{berndt1994using}, which measures the temporal alignment between generated and reference stroke sequences and reflects how closely the produced trajectories follow realistic writing motions.

For offline handwriting images, we compare our results against representative state-of-the-art models including One-DM~\cite{dai2024one}, DiffusionPen~\cite{nikolaidou2024diffusionpen}, VATr++~\cite{vanherle2024vatr++}, and Emuru~\cite{pippi2025zero}. 
We report \textbf{Fréchet Inception Distance (FID)}~\cite{heusel2017gans} and \textbf{Binarized FID (BFID)} to measure the distributional similarity between generated and real handwriting images. 
FID evaluates global visual realism, while BFID focuses on structural similarity after binarization, making it more sensitive to stroke shapes and character structures.

To measure handwriting style similarity, we use \textbf{Handwriting Distance (HWD)}~\cite{pippi2023hwd}, which quantifies stylistic differences between generated and reference handwriting samples. 
For readability evaluation, we compute \textbf{Character Error Rate (CER)} using the TrOCR recognizer~\cite{li2023trocr}, indicating how accurately the generated text can be recognized by an OCR system.

We additionally report \textbf{Kernel Inception Distance (KID)}~\cite{binkowski2018demystifying}, \textbf{Geometry Score (GS)}~\cite{khrulkov2018geometry}, and paired metrics including \textbf{Root Mean Square Error (RMSE)} and \textbf{Learned Perceptual Image Patch Similarity (LPIPS)}~\cite{zhang2018unreasonable} when ground truth correspondence is available. 
These additional metrics are reported in the extended quantitative results presented in the following section~\ref{sec:quali_off_extended}.

\subsection{Extended Quantitative Analysis}
\label{sec:quali_off_extended}
\begin{table}[tb]
\caption{Additional quantitative evaluation for baseline models and our framework on IAM and CVL datasets. \textbf{Bold}: Best/\underline{Underline}: Second-best.}
\vspace{-2mm}
\centering\small
\resizebox{0.9\linewidth}{!}{
\begin{tabular}{lcccccccc}
\toprule
\multicolumn{1}{c}{\textbf{Dataset}}
& \multicolumn{4}{c}{\textbf{IAM Words}}
& \multicolumn{4}{c}{\textbf{CVL Words}} \\
\cmidrule(lr){1-1} \cmidrule(lr){2-5} \cmidrule(lr){6-9}
\multicolumn{1}{c}{\textbf{Model}}
& KID$\downarrow$ & BKID$\downarrow$ & GS$\downarrow$ & RMSE$\downarrow$
& KID$\downarrow$ & BKID$\downarrow$ & GS$\downarrow$ & RMSE$\downarrow$\\
\midrule
\textbf{VATr++}
    & 0.0226 & 0.0221 & 0.0170 & 67.31 
    & \underline{0.0152} & 0.0144 & \textbf{0.0019} & \underline{56.14} \\
\textbf{DiffPen}
    & 0.1842 & 0.1703 & \textbf{0.0031} & 65.26 
    & 0.1186 & 0.1146 & 0.1574 & 60.65 \\
\textbf{One-DM}
    & \underline{0.0198} & \underline{0.0059} & 0.0073 & 79.61 
    & 0.0234 & \underline{0.0091} & 0.0110 & 67.82 \\
\textbf{Emuru}
    & 0.0857 & 0.0614 & 0.0047 & \underline{61.36} 
    & 0.0685 & 0.0484 & 0.2881 & 56.32 \\
\midrule
\textbf{DiffPen+Ours}
    & 0.0957 & 0.0868 & \underline{0.0041} & \textbf{22.14} 
    & 0.0477 & 0.0573 & 0.1549 & 59.61 \\
\textbf{One-DM+Ours}
    & \textbf{0.0133} & \textbf{0.0024} & 0.0062 & 78.55 
    & \textbf{0.0101} & \textbf{0.0059} & \underline{0.0020} & \textbf{39.19} \\
\bottomrule
\end{tabular}
}
\label{tab:offline_eval_ext}
\end{table}

Table~\ref{tab:offline_eval_ext} reports additional quantitative results on the IAM and CVL datasets. Our method consistently improves distributional metrics such as KID and BKID when integrated with diffusion-based generators. In particular, One-DM+Ours achieves the best KID and BKID scores on both datasets, indicating that our renderer produces handwriting samples that better match the real data distribution. Furthermore, our approach reduces RMSE when ground-truth correspondence is available, demonstrating improved pixel-level fidelity between generated and reference samples. Importantly, these improvements are achieved while maintaining comparable geometric consistency as measured by the Geometry Score (GS).

\subsection{User Study Design and Analysis}
\label{sec:user_study}

\begin{table}[t]
\centering\scriptsize
\caption{User study results (Lower is better).}
\label{tab:user_study}
\vspace{-2mm}
\resizebox{0.9\linewidth}{!}{%
\begin{tabular}{l c c c c c c}
\toprule
\multirow{2}{*}{\textbf{Model}} & 
\multirow{2}{*}{\textbf{VATr++}} & 
\multirow{2}{*}{\textbf{DiffPen}} & 
\multirow{2}{*}{\textbf{One-DM}} & 
\multirow{2}{*}{\textbf{Emuru}} & 
\textbf{DiffPen} & \textbf{One-DM} \\
& & & & & \textbf{+Ours} & \textbf{+Ours} \\ 
\midrule
Score $\downarrow$ & 3.11 & 3.18 & 3.40 & 5.51 & \underline{3.10} & \textbf{2.84} \\ 
\bottomrule
\end{tabular}%
}
\end{table}

To further evaluate the perceptual quality and fidelity of our method, we conduct a user study comparing our approach with existing state-of-the-art models.

\noindent\textbf{Experimental Setup.}
We recruited 15 participants via Amazon Mechanical Turk (MTurk) following the protocol of~\cite{shin2025video, Shin2024bw, park2024kinetic, jeon2026motion, jeon2026rebalancing, lee2026universal}. 
We selected 10 representative target words exhibiting diverse handwriting styles. 
For each test case, participants were presented with a target reference image and a set of generated samples produced by different models: VATr++, DiffPen, One-DM, Emuru, and our augmented variants (DiffPen+Ours and One-DM+Ours). 
Participants were asked to rank the generated results according to their visual similarity to the reference image, considering both stroke geometry and stylistic consistency.

\noindent\textbf{Results and Analysis.}
Table~\ref{tab:user_study} summarizes the results. 
We report the average ranking score, where lower values indicate higher preference and closer similarity to the target reference.

Our approach consistently improves the perceptual quality of diffusion-based handwriting generators. 
In particular, \textbf{One-DM+Ours} achieves the best score of 2.840, compared with 3.406 for the original One-DM. 
Similarly, DiffPen+Ours improves over DiffPen, from 3.180 to 3.106.
These results indicate that incorporating physically grounded stroke trajectories and brush parameters provides stronger structural guidance, enabling diffusion models to produce more realistic and stylistically faithful handwriting.

\subsection{Extended Qualitative Results}

\begin{figure}[!t]
\centering

\setlength{\abovecaptionskip}{2pt}
\setlength{\belowcaptionskip}{4pt}

\includegraphics[
    width=\linewidth,
    keepaspectratio
]{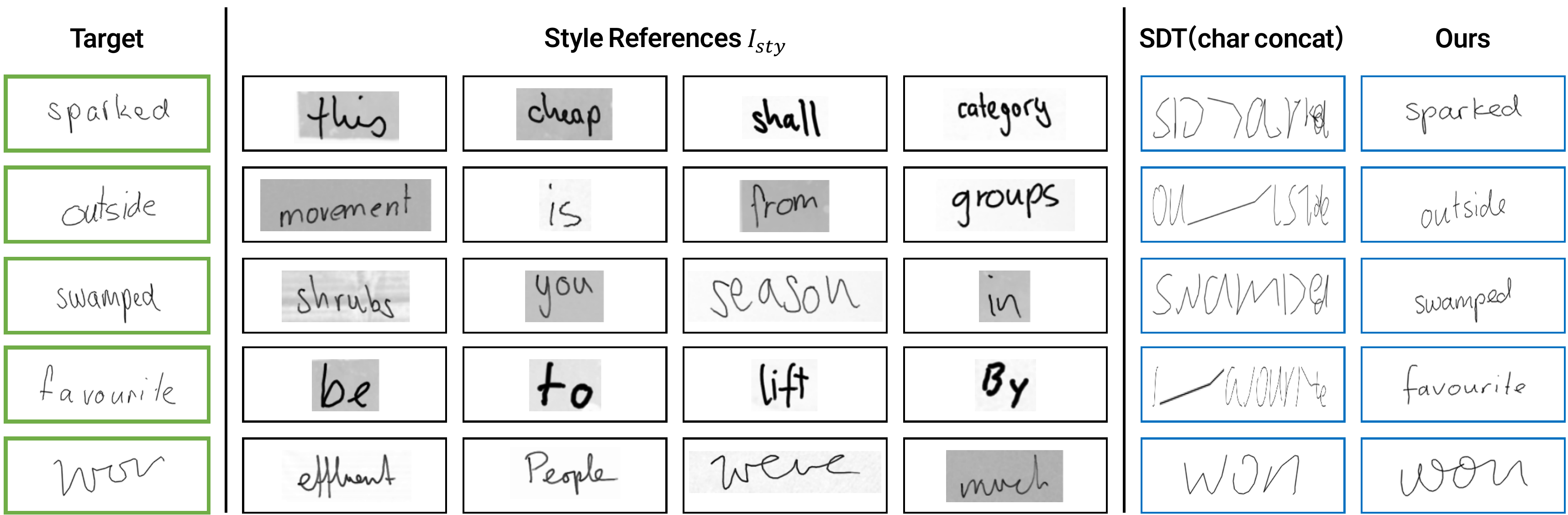}
\caption{Qualitative comparison of online handwriting generation results.}
\label{fig:quali_online}

\vspace{1mm}

\includegraphics[
    width=\linewidth,
    keepaspectratio
]{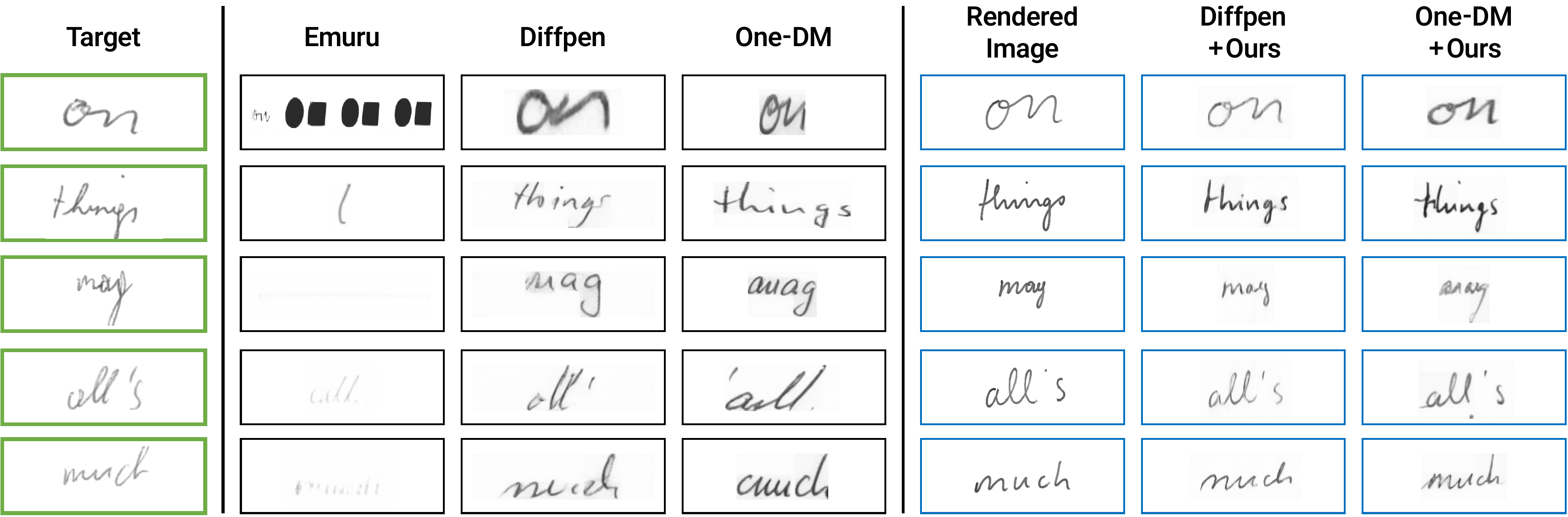}
\caption{Qualitative comparison of offline handwriting generation results
on the CVL dataset.}
\label{fig:quali_off_cvl}

\vspace{1mm}

\includegraphics[
    width=\linewidth,
    keepaspectratio
]{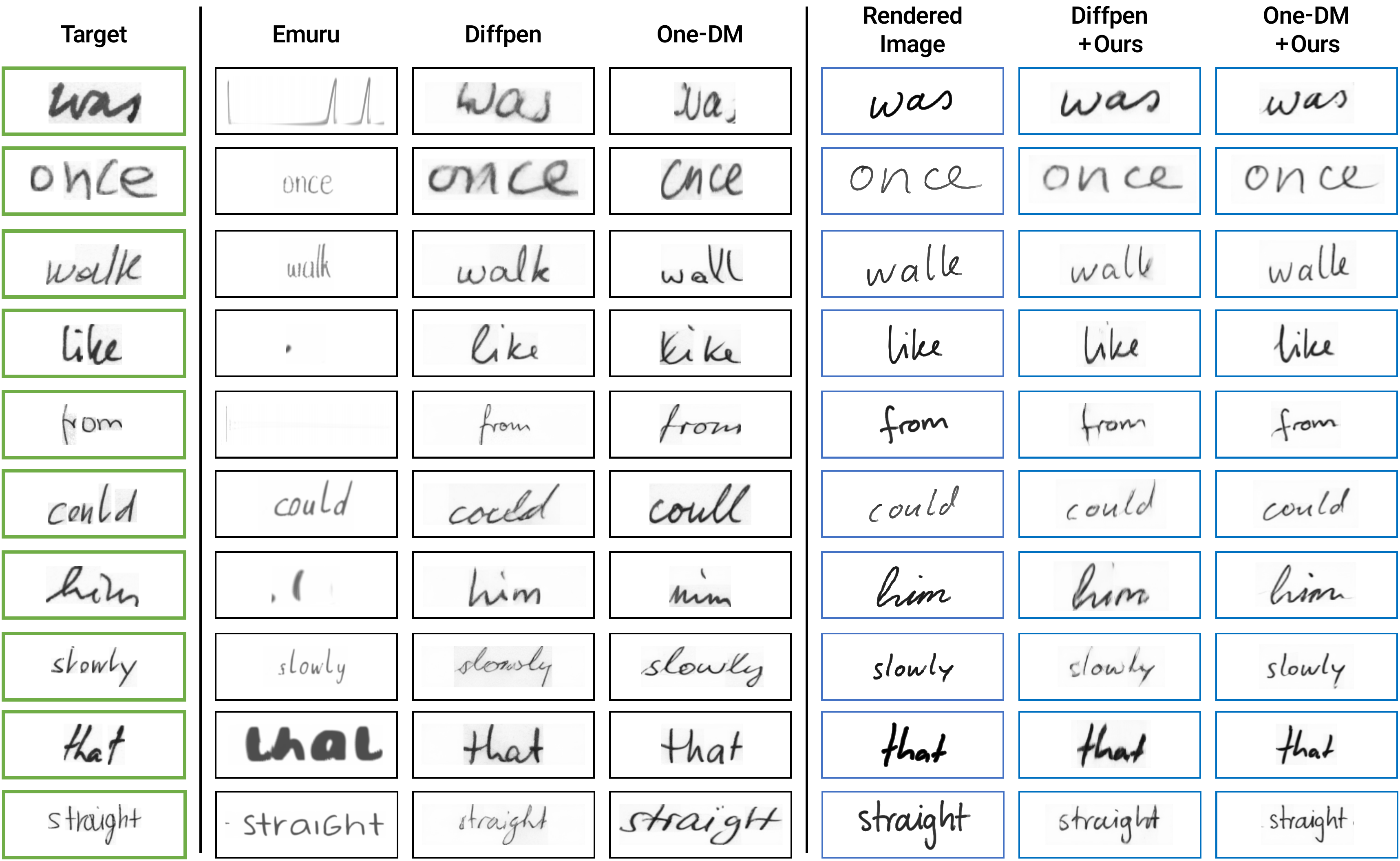}
\caption{Additional qualitative comparison of offline handwriting
generation results on the IAM dataset.}
\label{fig:quali_off_additive}

\end{figure}

The handwriting trajectory generation results of the stroke generator are shown in Fig.~\ref{fig:quali_online}. For SDT~\cite{dai2023disentangling}, which is a single-character generation model, characters are concatenated with fixed spacing to form words. 
Such single-character models require an additional layout step to arrange characters into words. In contrast, our model directly supports multi-character generation, allowing it to naturally handle connected cursive handwriting where letters are written continuously.

Fig.~\ref{fig:quali_off_cvl} presents qualitative results on the CVL dataset, while Fig.~\ref{fig:quali_off_additive} provides additional qualitative results on the IAM dataset beyond those shown in the main paper. Even without training on these offline handwriting datasets, the rendered images generated from predicted strokes and brush parameters exhibit structures that closely resemble real handwriting. Furthermore, when integrated with offline handwriting diffusion models, our approach produces handwriting images that are more realistic, effectively reducing the domain gap between rendered stroke structures and real handwriting images.

\clearpage

\clearpage
\bibliographystyle{splncs04}
\bibliography{main}
\end{document}